\documentclass[10pt,twocolumn,letterpaper]{article}

\usepackage{cvpr}
\usepackage{times}
\usepackage{epsfig}
\usepackage{graphicx}
\usepackage{amsmath}
\usepackage{amssymb}
\usepackage[percent]{overpic}
\usepackage{boldline}
\usepackage{multirow,enumitem}
\usepackage[numbers,sort,compress]{natbib}
\usepackage{booktabs}
\usepackage{multirow} 
\usepackage{xcolor}

\definecolor{mli_color1}{RGB}{102, 102, 255}
\definecolor{mli_color2}{RGB}{255, 117, 26}

\newcommand{\best}[1]{\textcolor{black}{\textbf{#1}}}

\usepackage{soul}
\usepackage{xcolor}

\usepackage{amsmath}
\usepackage{amsfonts}
\usepackage{amssymb}
\usepackage{amsthm}
\usepackage{bm}
\usepackage{bbm}
\usepackage{mathtools}
\usepackage{enumitem}
\usepackage{thmtools,thm-restate}
\usepackage[ruled,vlined]{algorithm2e}
\SetAlFnt{\small}
\usepackage{color}
\usepackage{graphicx}
\usepackage{comment}
\theoremstyle{plain}

\theoremstyle{definition}

\theoremstyle{remark}

\def\Ab{\mathbf{A}}\def\Bb{\mathbf{B}}

\def\Wb{\mathbf{W}}

\def\R{\Rbb}

\usepackage{etex,etoolbox}
\usepackage{environ}

\makeatletter
\providecommand{\@fourthoffour}[4]{#4}
\newcommand\fixstatement[2][\proofname\space of]{%
	\ifcsname thmt@original@#2\endcsname
	\AtEndEnvironment{#2}{%
		\xdef\pat@label{\expandafter\expandafter\expandafter
			\@fourthoffour\csname thmt@original@#2\endcsname\space\@currentlabel}%
		\xdef\pat@proofof{\@nameuse{pat@proofof@#2}}%
	}%
	\else
	\AtEndEnvironment{#2}{%
		\xdef\pat@label{\expandafter\expandafter\expandafter
			\@fourthoffour\csname #1\endcsname\space\@currentlabel}%
		\xdef\pat@proofof{\@nameuse{pat@proofof@#2}}%
	}%
	\fi
	\@namedef{pat@proofof@#2}{#1}%
}

\newcounter{proofcount}

\NewEnviron{proofatend}{%
	\edef\next{%
		\noexpand\begin{proof}[\pat@proofof\space\pat@label]%
			\unexpanded\expandafter{\BODY}}%
		\global\toks\numexpr\prooftoks+\value{proofcount}\relax=\expandafter{\next\end{proof}}
	\stepcounter{proofcount}}

\def\printproofs{%
	\count@=\z@
	\loop
	\the\toks\numexpr\prooftoks+\count@\relax
	\ifnum\count@<\value{proofcount}%
	\advance\count@\@ne
	\repeat}
\makeatother

\fixstatement{lemma}
\fixstatement{theorem}
\fixstatement{proposition}
\fixstatement{corollary}

\def\R{\mathbb R}

\makeatletter
\newcommand\notsotiny{\@setfontsize\notsotiny\@vipt\@viipt}
\makeatother

\newcommand{\SKIP}[1]{}

\usepackage[pagebackref=true,breaklinks=true,letterpaper=true,colorlinks,bookmarks=false]{hyperref}

\cvprfinalcopy % *** Uncomment this line for the final submission

\def\cvprPaperID{9531} % *** Enter the CVPR Paper ID here

\ifcvprfinal\fi
\begin{document}
%%%%%%%%% TITLE

\title{MMTM: Multimodal Transfer Module for CNN Fusion}

\author{
Hamid Reza Vaezi Joze\thanks{Equal contribution.}\\
Microsoft\\
%One Microsoft way, Redmond, WA 98052, USA\\
{\tt\scriptsize hava@microsoft.com}
\and
Amirreza Shaban$^*$\thanks{Work done during an internship at Microsoft.}\\
Georgia Tech \\
{\tt\scriptsize{amirreza@gatech.edu}}
\and
Michael L. Iuzzolino$^\dagger$\\
CU Boulder\\
{\tt\notsotiny michael.iuzzolino@colorado.edu }
\and
Kazuhito Koishida\\
Microsoft\\
{\tt\scriptsize kazukoi@microsoft.com}
}

\maketitle
\thispagestyle{empty}

\begin{abstract}
In late fusion, each modality is processed in a separate unimodal Convolutional Neural Network (CNN) stream and the scores of each modality are fused at the end. Due to its simplicity, late fusion is still the predominant approach in many state-of-the-art multimodal applications. In this paper, we present a simple neural network module for leveraging the knowledge from multiple modalities in convolutional neural networks. The proposed unit, named Multimodal Transfer Module (MMTM), can be added at different levels of the feature hierarchy, enabling slow modality fusion. Using squeeze and excitation operations, MMTM utilizes the knowledge of multiple modalities to recalibrate the channel-wise features in each CNN stream. Unlike other intermediate fusion methods, the proposed module could be used for feature modality fusion in convolution layers with different spatial dimensions. Another advantage of the proposed method is that it could be added among unimodal branches with minimum changes in the their network architectures, allowing each branch to be initialized with existing pretrained weights. Experimental results show that our framework improves the recognition accuracy of well-known multimodal networks. We demonstrate state-of-the-art or competitive performance on four datasets that span the task domains of dynamic hand gesture recognition, speech enhancement, and action recognition with RGB and body joints.
\end{abstract}

%-------------------------------------------------------------------------
\section{Introduction}
Different sensors can provide complementary information about the same context. Multimodal fusion is the act of extracting and combining relevant information from the different modalities that leads to improved performance over using only one modality. 
This technique is widely used in various machine learning tasks, such as video classification~\cite{karpathy2014large,yang2016multilayer}, action recognition~\cite{owens2018audio}, emotion recognition~\cite{kim2013deep,liu2016emotion}, and audio visual speech enhancement~\cite{afouras2018conversation, ephrat2018looking}.

\begin{figure}[t]
    \vspace{-.1in}
	\begin{center}
		%\fbox{\includegraphics[width=\linewidth]{focal_loss.pdf}}
		\includegraphics[width=.9\linewidth]{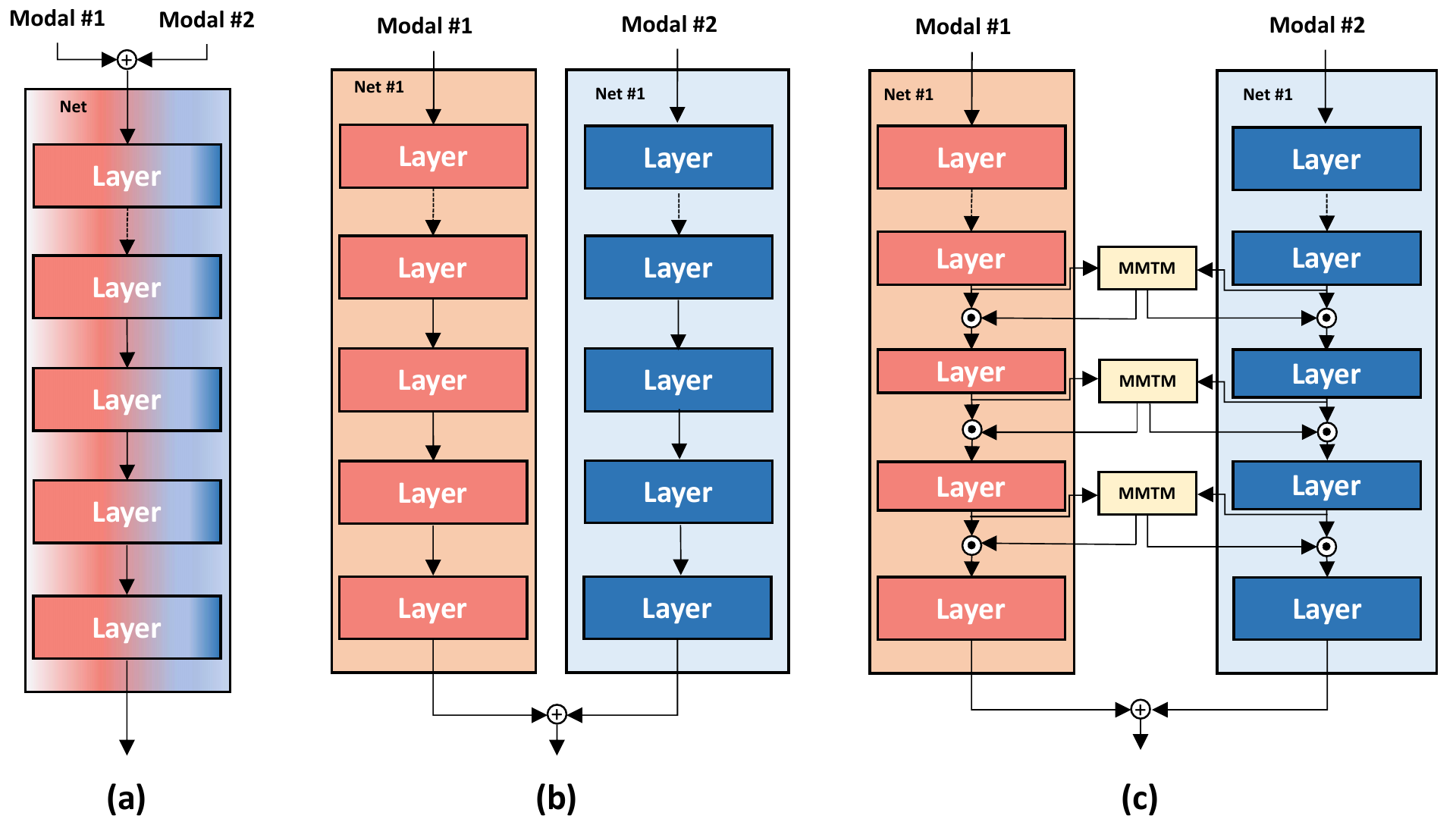}
	\end{center}
	\vspace{-.15in}
	\caption{(a) early fusion (b) late fusion (c) intermediate fusion with Multimodal Transfer Module (MMTM). MMTM operates between CNN streams and uses information from different modalities to recalibrate channel-wise features in each modality.} % It leaves the main architecture of each branch intact allowing knowledge transfer for each stream using the pretrained network with their unimodal training data.}
	\label{fig:fusion}
	\vspace{-0.2in}
\end{figure}

In general, fusion can be achieved at the input level (i.e. early fusion), decision level (i.e. late fusion), or intermediately~\cite{ramachandram2017deep}. Although studies in neuroscience~\cite{schroeder2005multisensory,macaluso2006multisensory} and machine learning~\cite{owens2018audio,karpathy2014large} suggest that mid-level feature fusion could benefit learning, late fusion is still the predominant method utilized for mulitmodal learning~\cite{liu2018recognizing,abavisani2019unimodal,katsaggelos2015audiovisual}. This is mostly due to practical reasons. For example, a simple pooling operator~\cite{simonyan2014two,natarajan2012multimodal} or an attention mechanism~\cite{jacobs1991adaptive} can be used to fuse 1-dimensional prediction scores of each stream. However, intermediate level features of different modalities have different or unaligned spatial dimensions making the intermediate fusing more challenging. 
Another reason for the popularity of late fusion is that the architecture of each unimodal stream is carefully designed over years to achieve state-of-the-art performance for each modality. This also enables the CNN streams of a multimodal framework to be initialized by weights that have been pretrained with a large number of unimodal training samples. However, intermediate fusion requires major changes in the base network architecture, which complicates the use of pretrained weights in most cases and requires the network to be retrained from randomly initialized states~\cite{perez2019mfas,vielzeuf2018centralnet}. Figure~\ref{fig:fusion} illustrates three common multimodal fusion techniques.

The goal of the proposed method is to overcome the aforementioned problems of intermediate fusion. Inspired by the squeeze and excitation (SE) module~\cite{Hu_2018_CVPR} for unimodal convolutional neural networks, we propose a multimodal transfer module to recalibrate the channel-wise features of different CNN streams. 
MMTMs can be inserted into intermediate levels of any late fusion backbone architecture. Each MMTM has two units: a) a multimodal squeeze unit that receives the features from all modalities at a given level of representation across the branches, generating a global joint representation of these features, and b) an excitation unit that uses this joint representation to adaptively emphasize on more important features and suppress less important ones in all modalities. The squeeze unit aggregates spatial dimensions, allowing information with global receptive fields from all modalities to be used in the global representation. It also enables learning a joint representation from modalities with different spatial dimensions.

Although the module design is generic and could potentially be added at any level in the network hierarchy, the optimal locations and number of modules are different for each application. We design application specific networks for gesture recognition, audio-visual speech enhancement, and action recognition tasks and study the benefit of adding MMTM in their architectures. We make the following empirical observations from these applications. Firstly, adding MMTM to intermediate and high-level features is beneficial, whereas the same is not true about low-level features. We believe that is because intera-modality correlation in low-level features is lower compared to intermediate and high-level features. This is also highlighted in previous research~\cite{li2017modout}. Secondly, even in gesture recognition where RGB and depth modalities are spatially aligned and fusion can be done without the squeeze operation, squeezing considerably improves the performance by providing information with a global receptive field. Lastly, excitation by gating operation outperforms the sum operation that is usually used in residual learning, highlighting the importance of the emphasis and suppression mechanisms.

In summary, this paper makes the following contributions: First, we propose a new neural network module called MMTM to fuse knowledge from intermediate features of unimodal CNNs. Second, we design different network architectures for three different multimodal fusion applications: gesture recognition using multiple visual modalities, audio-visual speech enhancement, and action recognition with RGB and body joints. We demonstrate through experiment on these tasks that MMTM improves the performance beyond the late fusion approach.
\section{Related Work}
\label{sec:relatedwork}
% ********************************
% BACKUP of original is here
% ------------------------------------
% \input{backups/section_2.tex}
% ------------------------------------
% ********************************
%In general, multimodal fusion can be achieved at the feature level (i.e. early fusion), decision level (i.e. late fusion) or intermediately~\cite{ramachandram2017deep}. 
In late fusion, the prediction of each unimodal stream are fused to make the final prediction. Fusion can be via element-wise summation, a weighted average~\cite{natarajan2012multimodal}, a bilinear product~\cite{ben2017mutan}, or a more sophisticated rank minimization~\cite{ye2012robust} method. Another approach to late fusion utilizes attention to pick the best expert for each input signal~\cite{jacobs1991adaptive}. The gated multimodal units~\cite{arevalo2017gated} extends this method by enabling gating at intermediate feature levels. More recently, Hu \etal~\cite{hu2019dense} propose a dense multimodal intermediate fusion network for hierarchical joint feature learning. Similar to~\cite{arevalo2017gated}, the dense fusion operator in~\cite{hu2019dense} assumes identical spatial dimensions for different streams. Despite the similarity of these approaches to our work, their applicability is limited to layers where the multimodal features' spatial dimensions are the same, or at the very end of the network where spatial dimensions are already aggregated. The squeeze operation proposed in this work allows the fusion of modalities with different spatial dimensions at any level of the feature hierarchy.

In a related multimodal learning topic, called cross-modal learning, information from multiple modalities are used to improve the learning performance within any individual modality. It is assumed that data from all the modalities are present during training but the performance is tested on only one modality~\cite{ngiam2011multimodal}. MTUT~\cite{abavisani2019unimodal} uses spatiotemporal semantic alignment loss to improve the performance of each stream in gesture recognition. We believe cross-modal learning approaches are orthogonal to our work since the improved unimodal networks learned by these methods can initialize weights of the CNN streams in our model.

{\bf \noindent Multimodal Action Recognition in Videos} Video~\cite{simonyan2014two,karpathy2014large, khodabandeh2018diy} and skeleton~\cite{zhang2019view, liu2018recognizing,luvizon20182d} modalities have been extensively used for the action recognition task. Each of these approaches have their own drawbacks. With the lack of explicit human body model, video based action recognition methods deal poorly with background clutter and non-action movements~\cite{liu2018recognizing}. On the other hand, by solely relying on body pose most of contextual and global cues present in the video will be lost. Recent methods develop architectures to fuse these modalities to further improve the performance of action recognition. In~\cite{luvizon20182d}, an end-to-end trainable multitask network for joint pose estimation and action recognition is proposed. PoseMap~\cite{liu2018recognizing} utilizes a two stream network to process spatiotemporal pose heatmaps and skeleton separately, and uses late fusion for the final prediction. A bilinear pooling block that separately pools input features in modality and time directions is employed in~\cite{Hu_2018_ECCV}.

{\bf \noindent Audio-Visual Speech Enhancement (AVSE)} Work in AVSE is strongly motivated by the cocktail party effect, which refers to humans' ability to selectively attend to auditory signals of interest within a noisy environment. Experiments in neuroscience have demonstrated that cross-modal integration of audio-visual signals may improve the perceptual quality of the targeted acoustic signal~\cite{ghazanfar2003neuroperception, partan1999communication, rowe2002sound}. Inspired by the results from biological research, recent studies focus on augmenting audio only speech enhancement methods with visual information, such as lip movement. State-of-the-art results have been achieved by recent AVSE models that use deep neural networks~\cite{afouras2018conversation, ephrat2018looking, afouras2019my, hou2018audio}. The predominant approach taken for AV fusion is late fusion~\cite{katsaggelos2015audiovisual}, where the audio and visual information is processed separately then integrated at a singular point via channel-wise concatenation. 

{\bf \noindent Hand Gesture Recognition} Interpreting hand gestures via machine learning algorithms is significantly important in human-computer interaction. 
We review the 3D convolutional neural network based gesture recognition algorithms~\cite{molchanov2015hand,molchanov2015multi,molchanov2016online,camgoz2016using,miao2017multimodal} that are designed for processing time series data among other branches~\cite{zhang2017learning,cao2017egocentric,cui2017recurrent,zhu2017multimodal}. In~\cite{molchanov2015hand}, a novel 3D CNN is proposed to integrate depth and image gradient values to recognize dynamic hand gestures. Molchanov \etal~\cite{molchanov2015multi} employ a multistream 3D CNN to fuse streams of data from multiple sensors including short-range radar, color, and depth sensors for recognition. A real-time method is presented in~\cite{molchanov2016online} to simultaneously detect and classify gestures in videos. Camgoz \etal~\cite{camgoz2016using} present a late fusion approach for fusing the scores of unimodal 3D CNN streams. Miao \etal propose ResC3D~\cite{miao2017multimodal}, a 3D CNN architecture that combines multimodal data using an attention model. MFFs~\cite{Kopuklu_2018_CVPR_Workshops} develops a data level fusion method for RGB and optical flow. FOANet~\cite{Narayana_2018_CVPR} proposes a sparse fusion technique for hand gesture recognition. FOANet decomposes each input modality (RGB, depth, and $2$ types of optical flow) into separate focus channels (global, right hand, left hand) and processes each of these $12$ focus channels in an independent unimodal network. Finally, it learns a sparsely connected late fusion network to avoid overfitting. Unlike our method, FOANet relies on the output of a detector to find the focus areas in the video.
%In an alternative approach, some CNN-based hand gesture recognition models have utilized recurrent architectures to capture the temporal information~\cite{zhang2017learning,cao2017egocentric,cui2017recurrent,zhu2017multimodal}.

{\bf \noindent Squeeze and Excitation (SE)
Network~\cite{Hu_2018_CVPR}} Our proposed method can be seen as a generalization to the SE module, which is proposed for unimodal deep neural networks. The SE modules uses self excitation to adaptively recalibrate channel-wise feature responses. Our work adopts the SE module for multimodal feature recalibrations.

\section{Multimodal Transfer Module}
\label{sec:proposed}
\begin{figure}[t]
	\vspace{-0.1in}
	\begin{center}
		%\fbox{\includegraphics[width=\linewidth]{focal_loss.pdf}}
		\includegraphics[width=.78\linewidth]{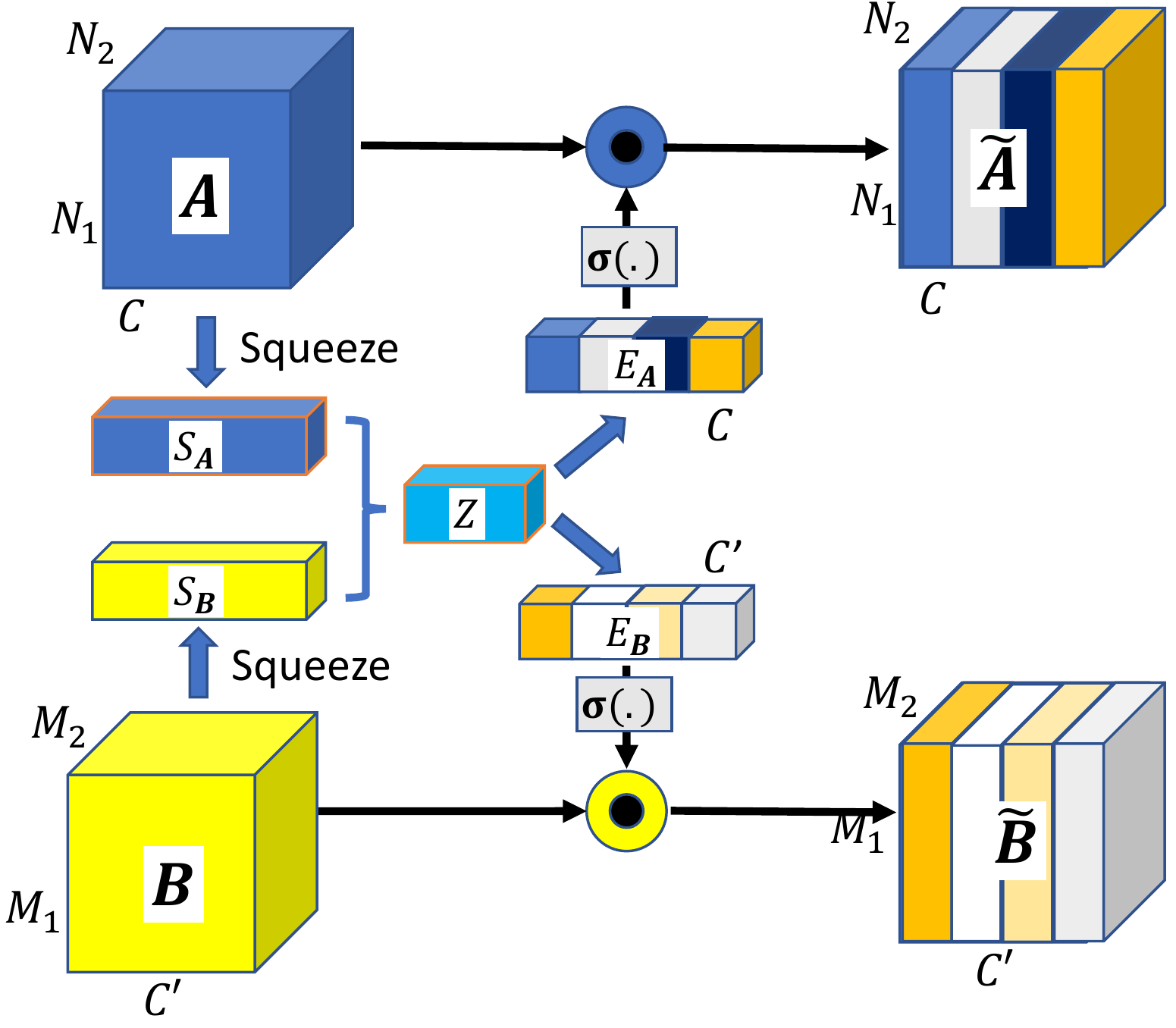}
	\end{center}
		\vspace{-0.1in}
	\caption{Architecture of MMTM for two modalities. $\Ab$ and $\Bb$, that represent the features at a given layer of two unimodal CNNs, are the inputs to the module. For better visualization we limit the number of their spatial dimensions to $2$. MMTM uses squeeze operations to generate global feature descriptor from each tensor. Both tensors are map into a joint representation $Z$ by using concatenation and fully-connected layer. The excitation signals $E_\Ab$ and $E_\Bb$ are generated using the joint representation. Finally the excitation signals are used to gate the channel-wise features in each modality.}
	\label{fig:mmtl}
\end{figure}

% ********************************
% BACKUP of original is here
% ------------------------------------
% \input{backups/section_3.tex}
% ------------------------------------
% ********************************
%\amir{Agree but run of out time :(} \mlic{Still seeking clear way to introduce this section}
% Let CNN$_i$ denote CNN stream $i$ that processes a single modality of the input data. 
In this section, we discuss the simplest case of fusion between two disjoint CNN streams, CNN$_1$ and CNN$_2$. %Note that our approach generalizes to arbitrarily many modalities. 
Let $\Ab \in \R^{N_1\times \dots \times N_K \times C}$ and $\Bb \in \R^{M_1\times \dots \times M_L \times C'}$ represent the features at a given layer of CNN$_1$ and CNN$_2$, respectively. Here, $N_i$ and $M_i$ represent the spatial dimensions\footnote{In general, it is possible to have more than two (e.g. time dimension in 3D convolutions could be treated as a spatial dimension) or no (e.g. fully connected layers) spatial dimensions.}, and $C$ and $C'$ represent the number of channels of the corresponding features in CNN$_1$ and CNN$_2$ respectively. MMTM receives features $\Ab$ and $\Bb$ as input, learns a global multimodal embedding from them, and uses this embedding to recalibrate the input features. This is done in a two-step multimodal squeeze and excitation process described below.

{\bf \noindent Squeeze} The information in the output features of convolution layers are limited by the size of their receptive fields and lacks global context. As suggested by~\cite{Hu_2018_CVPR}, we first squeeze the spatial information into the channel descriptors via a global average pooling over  spatial dimensions of the input features:
\begin{align}
    S_\Ab(c) &= \frac{1}{\prod_{i=1}^{K} N_i}\sum_{n_1,\dots, n_K} \Ab(n_1, \dots, n_K, c) \\
    S_\Bb(c) &= \frac{1}{\prod_{i=1}^{L} M_i}\sum_{m_1,\dots, m_L} \Bb(m_1, \dots, m_L, c).
\end{align}
Importantly, the squeeze operation enables fusion between modalities with features of arbitrary spatial dimension. Note that while we use simple average pooling, more sophisticated pooling methods could be used at this step.

{\bf \noindent Multimodal Excitation} The function of this unit is to generate the excitation signals, $E_\Ab \in \R^C$ and $E_\Bb \in \R^{C'}$, which can be used to recalibrate the input features, $\Ab$ and $\Bb$, by a simple gating mechanism:
\begin{align*}
    \tilde{\Ab} &= 2 \times \sigma(E_\Ab) \odot \Ab \\
    \tilde{\Bb} &= 2 \times \sigma(E_\Bb) \odot \Bb, 
\end{align*}
where $\sigma(.)$ is the sigmoid function and $\odot$ is the channel-wise product operation. This allows the suppression or excitation of different filters in each stream. Note that the MMTM weights are regularized in order to control the proximity of $E_\Ab$ and $E_\Bb$ to zero. Specifically, increasing the regularization weight of $E_\Ab$ pushes the gating signal $2 \times \sigma(E_\Ab)$ closer to the identity vector, limiting the effect of gating on feature $\Ab$.

The gating signals must apply different calibration weights to different modalities based on the same input representation. We achieve this by first predicting a joint representation $Z \in \R^{C_Z}$ from the squeezed signals
\begin{equation}
\label{eq:joint_fea}
Z = \Wb [S_\Ab, S_\Bb] + b,
\end{equation}
and then predicting excitation signals for each modality through two independent fully-connected layers
\begin{equation}
\label{eq:fc_excitation}
E_\Ab = \Wb_\Ab Z + b_\Ab,\quad E_\Bb = \Wb_\Bb Z + b_\Bb.
\end{equation}
Here, $[\cdot, \cdot]$ represents the concatenation operation, $\Wb \in \R^{C_Z \times (C+C')}, \Wb_\Ab \in \R^{C\times C_Z}, \Wb_\Bb \in \R^{C'\times C_Z}$ are the weights, and $b \in \R^{C_Z}, b_\Ab \in \R^C, b_\Bb \in \R^{C'}$ are the biases of the fully connected layers. As suggested in~\cite{Hu_2018_CVPR}, we use $C_Z = (C+C')/4$ to limit the model capacity and increase the generalization power. For fusing more than two modalities, we simply generalize this approach by concatenating squeezed features from all the modalities in Equation~\ref{eq:joint_fea} and predict excitation signals for each modality with an independent fully-connected layer like in Equation~\ref{eq:fc_excitation}.

Learning the joint representation in this way allows the features of one modality to recalibrate the features of another modality. For instance, in gesture recognition when a gesture is blurry in RGB camera and more apparent in depth modality, MMTM cross-modal recalibration affords more efficient processing in the RGB stream. 
Figure~\ref{fig:mmtl} summarizes the overall architecture of the proposed MMTM.

\section{Applications}
The MMTM is generic and can be easily integrated to any multimodal CNN architectures. In this section, we explore a few applications that can benefit from MMTM and describe the architecture  changes necessary to support multimodal fusion. We evaluate the performance of the proposed multimodal models in the experiment section.
\subsection{Hand Gesture Recognition}
\label{sec.app.visual}
Hand gesture recognition is a video classification task. It is shown that complementary sensory information, such as depth and optical flow, improves the performance of the gesture recognition~\cite{Kopuklu_2018_CVPR_Workshops,abavisani2019unimodal,cao2017egocentric,molchanov2016online}. There are multiple multimodal datasets available for this task~\cite{cao2017egocentric,zhang2018egogesture,molchanov2016online,joze2018ms} and several previous fusion methods have reported their results on these datasets~\cite{molchanov2015multi,molchanov2016online,camgoz2016using,miao2017multimodal}.

We design a gesture recognition network for fusing RGB, depth, and optical flow video streams via MMTM. To process the temporal inputs, we use I3D network architecture~\cite{carreira2017quo} with an inflated inception-v1~\cite{ioffe2015batch} backbone for all the streams. In I3D network, convolution and pooling kernels of the backbone network are expanded into 3D, enabling efficient spatial-temporal feature processing. %I3D network performs well on video classification tasks~\cite{carreira2017quo}. %Figure~\ref{fig:i3d_mmtl} shows designed architecture of a network design for fusing RGB, depth, and optical flow modalities. 
We apply MMTM after the last $6$ inception modules (the connectivity is similar to figure~\ref{fig:fusion}). %The input features to MMTM are $\Ab,\Bb \in \R^{t \times w \times h \times C}$, where $t$ represents temporal dimension and $w,h$ represent the spatial dimensions. 
Note that the output of 3D convolutions has a time dimension in addition to height, width, and channel dimensions. 
We empirically find that the best performance is achieved when the squeeze operation is applied over all the dimensions except for the channel dimension.
%
% Model Figure
\begin{figure}[htp]
\vspace{-.25in}
    \centering
    \includegraphics[width=0.75\columnwidth]{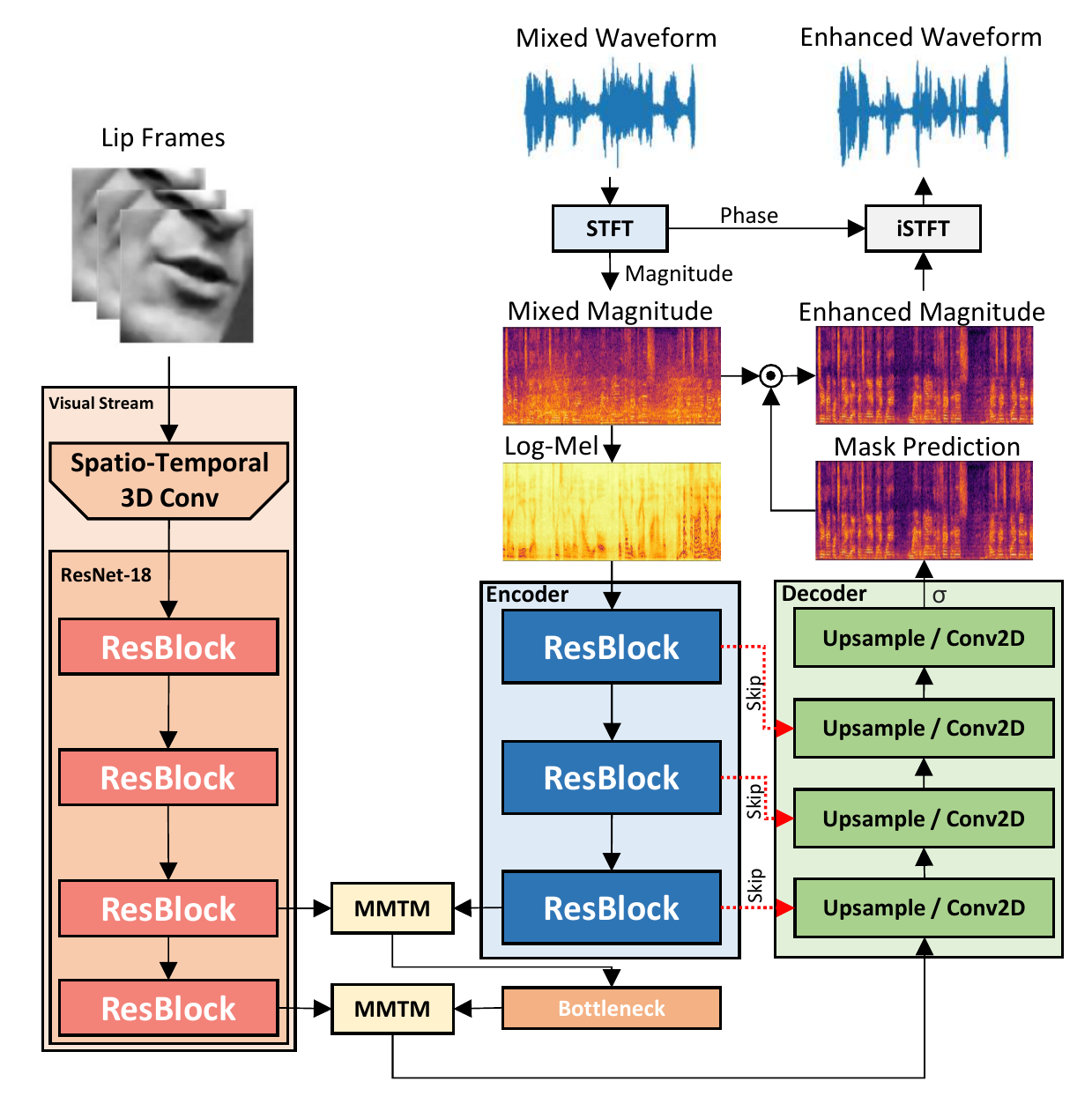}
    \caption{An overview of our AVSE  architecture.}
    \label{fig:avse_model}
\vspace{-.1in}
\end{figure}
\subsection{Audio-Visual Speech Enhancement}
The predominant method for AV speech enhancement combines audio and visual signals via channel-wise concatenation (CWC) using the late fusion approach. As an application of MMTM, we explore AV fusion for speech enhancement tasks using MMTM instead of the CWC-based late fusion. Model details are provided below, and an overview of our AVSE architecture can be found in Figure~\ref{fig:avse_model}. 

{\bf \noindent Visual Network} We use the spatio-temporal residual network proposed by \cite{stafylakis2017combining}, which consists of a 3D spatio-temporal convolution followed by a 2D ResNet-18 \cite{he2016deep}. Processing 3D features in a 2D convolution operation is achieved by packing the temporal dimension, $t$, into the batch dimension. The network is randomly initialized and trained concurrently with the AVSE task. 

{\bf \noindent Audio Network} Our audio network is an autoencoder with skip connections; we follow the design detailed in \cite{michelsanti2019training}. Figure \ref{fig:avse_model} (top) depicts the audio processing strategy, which follows the audio processing procedures of \cite{afouras2018conversation} and is detailed in Section~\ref{sec:av_experiment}. The network takes a log-mel mixture magnitude spectrogram, log-mel$(X_{mix})$, as input and outputs the predicted ideal ratio mask, $M$. The enhanced magnitude spectrogram, $X_{enh}$, is obtained via $X_{enh} = M \odot X_{mix}$, where $\odot$ denotes element-wise multiplication. The network is trained by minimizing the reconstruction loss between the enhanced magnitude, $X_{enh}$, and the target magnitude, $X_{spec}$, where $X_{spec}$ is obtained via short-time Fourier transform (STFT) from the target waveform. The optimization objective is given by $\mathcal{L} = ||X_{enh}  - X_{spec}||_1$.

{\bf \noindent Audio-Visual Fusion via MMTM} Let $F_a^j$ denote the audio feature at layer $j$ of the autoencoder with $F_a^j \in \mathbb{R}^{b \times t \times f \times c_a}$, where $b$, $t$, $f$, and $c_a$ are the batch, temporal, frequency, and audio channel dimensions, respectively. Let $F_v^i$ denote visual feature at layer $i$ of the visual network's ResNet-18 with $F_v^i \in \mathbb{R}^{b\cdot t \times h \times w \times c_v}$, where $h$, $w$ are spatial dimensions and $b, t, c_v$ are the batch, temporal, and visual channel dimensions, respectively. We unpack $t$ from the batch dimension of $F_v^i$ via reshaping such that $F_v^i \in \mathbb{R}^{b \times t \times h \times w \times c_v}$. The MMTM takes $F_a$ and $F_v$ as input and carries out the fusion procedure detailed in Section~\ref{sec:proposed}. For AVSE, the final output is from audio tower; consequently, MMTM does not gate on visual network.

\subsection{Human Action Recognition}
\label{sec:action_recognition_app}
Recent methods in human activity recognition %use late fusion approach to 
combine video and 3D pose information to further improve the performance of action recognition~\cite{liu2018recognizing,luvizon20182d,Hu_2018_ECCV}. Following the same approach, we utilize MMTM for intermediate fusion between a visual and a skeleton based network. Similar to the gesture recognition application, we use I3D for the RGB video stream and HCN, as suggested by~\cite{li2018co}, for the skeletal stream. Although HCN is not the sate-of-the-art for skeleton-based action recognition, the simplicity of its design makes it suitable for our approach. 

\begin{figure}
    \vspace{-.1in}
	\begin{center}
		\includegraphics[width=.9\linewidth]{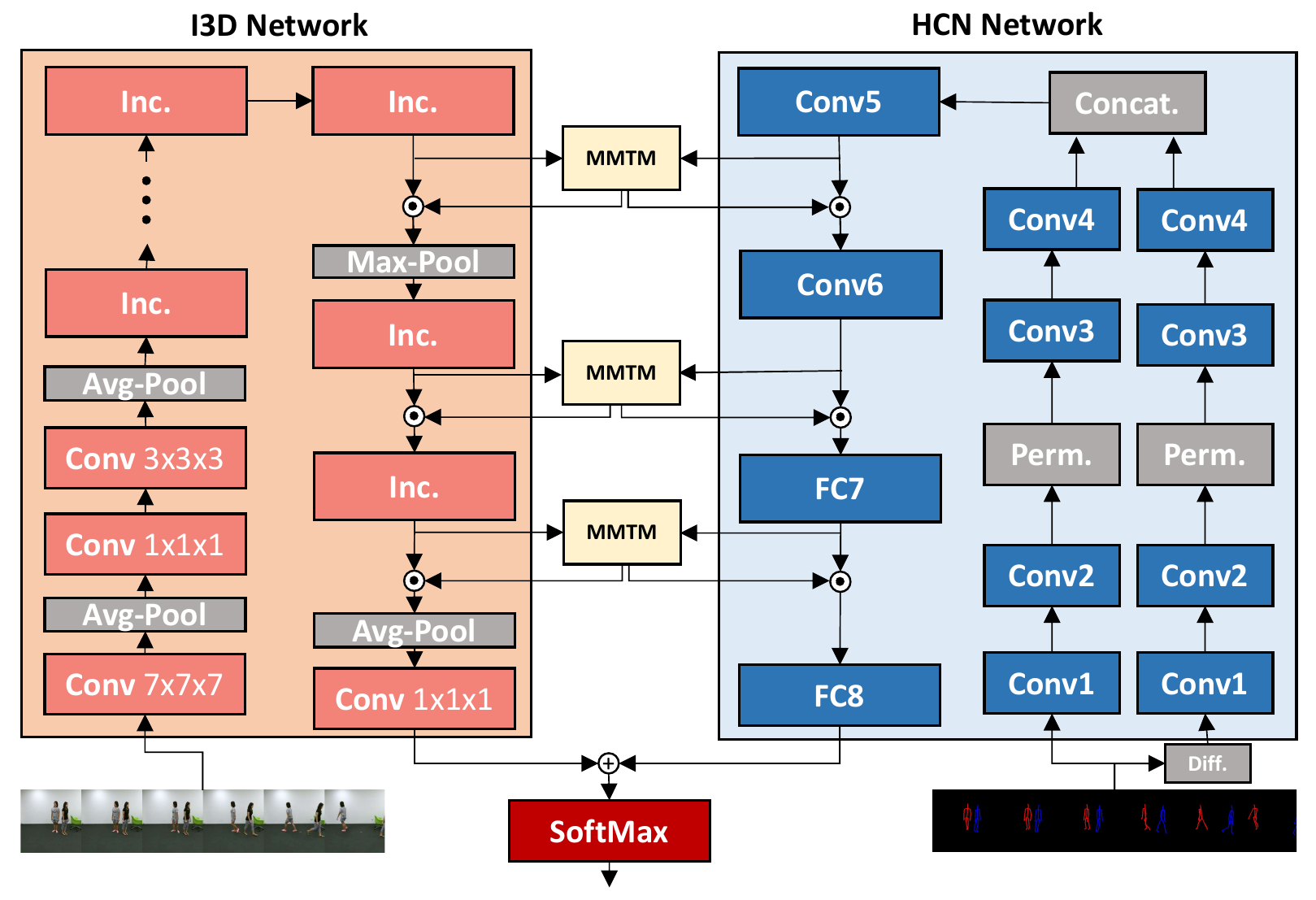}
	\end{center}
	\vspace{-.15in}
	\caption{Proposed multimodal architecture for action recognition. Each ``Inc.'' block represents an inception module described in~\cite{carreira2017quo}.}
	\label{fig:rgb_sk}
	\vspace{-0.15in}
\end{figure}

As it is shown in Figure~\ref{fig:rgb_sk}, HCN is comprised of two 2D convolution subnetworks: one branch processes the raw skeleton data, and the other branch processes the motion--the temporal gradients of the skeletal data. The two subnetworks are fused via channel-wise concatenation and followed by two convolution operations (conv5 and conv6), and finally, a fully connected layer (fc7).

Figure~\ref{fig:rgb_sk} illustrates the complete network we are proposing. We add $3$ MMTMs that receive inputs from last three inception modules of the I3D and conv5, conv6, and fc7 of HCN network. Let $\Ab \in \R^{t \times w \times h \times C}$ represent an I3D feature, where $t$ represents temporal dimension and $w,h$ are the spatial dimensions. Let $\Bb \in \R^{t \times n \times C'}$ represent HCN features after conv5 and conv6 layers,  where $t$ is the temporal dimension and $n$ is the body-joints dimension. The output of the fully connected layer (fc7) in HCN network is a 1-dimensional vector with no spatial dimension. In MMTMs, we aggregate all the dimensions of the inputs $\Ab$ and $\Bb$ except the channels. 
%Associating $\Ab$ and $\Bb$ would be inputs to MMTM as described in section~\ref{sec:proposed}.
The dimensions of the I3D and HCN features sent to the MMTMs ($\Ab$ and $\Bb$) do not match, but MMTM's squeezing operation makes the fusion possible.

\section{Experimental Results}\label{sec:result}
In this section, we evaluate the performance of the proposed method in gesture recognition, speech enhancement, and action recognition tasks. Due to the large number of experiments, we use a simple rule to decide the number of MMTMs in each architecture without an extensive architecture tuning scheme. We use MMTMs after each module in the second half of the network with minimum depth. This is $6$ MMTMs for hand gesture recognition experiments, $2$ in speech enhancement, and $3$ in action recognition experiment. Refer to Section~\ref{sec:study} for the study of the number of MMTMs in hand gesture recognition task.

\begin{figure*}[t]
\begin{center}
\includegraphics[width=0.9\linewidth]{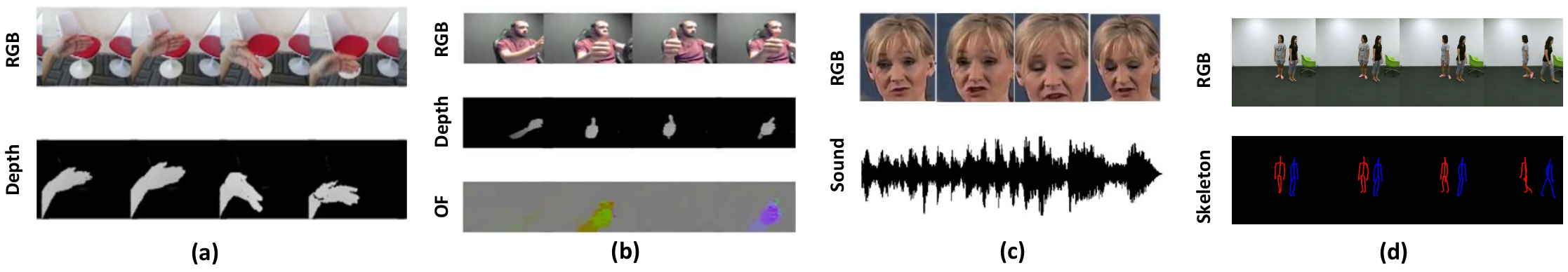}
\end{center}
\vspace{-0.15in}
   \caption{Sample sequences from multimodal datasets: (a) EgoGesture~\cite{cao2017egocentric} (b) NVGesture~\cite{molchanov2016online} (c) VoxCeleb2 \cite{chung2018voxceleb2} (d) NTU-RGBD~\cite{Shahroudy_2016_CVPR}}
\label{fig:datasets}
\vspace{-0.1in}
\end{figure*}

\subsection{Hand Gesture Recognition}
\label{sec:hand_gesture_result}
In this section, we evaluate our method against state-of-the-art dynamic hand gesture methods. We conduct experiments on two recent publicly available multimodal dynamic hand gesture datasets: \emph{EgoGesture}~\cite{cao2017egocentric,zhang2018egogesture} and  \emph{NVGestures}~\cite{molchanov2016online} datasets. Figure~\ref{fig:datasets} (a), (b) shows sample frames from the different modalities of these datasets.

{\bf \noindent Implementation Details:} In the design of our method, we adopt the architecture of I3D network~\cite{carreira2017quo} as the backbone network for each modality. The architecture details can be found in Section~\ref{sec.app.visual}. We start with the publicly available ImageNet~\cite{deng2009imagenet} + Kinetics~\cite{kay2017kinetics} pretrained networks for all of our experiments on I3D.
We optimize the objective function with the standard SGD optimizer using a momentum of $0.9$. We start with the base learning rate of $10^{-2}$ and reduce it $10\times$ when the loss is saturated. We use a batch size of $4$ containing $64$-frames ($32$-frames for \emph{EgoGesture}) snippets in the training stage. %The models were implemented in Tensor-Flow $1.9$~\cite{abadi2016tensorflow}. 
%Similar to~\cite{carreira2017quo}
We employ the following spatial and temporal data augmentations during the training stage. For spatial augmentation, videos are resized to $256\times256$ pixels, and then randomly cropped with a $224\times224$ patch. The resulting video is randomly flipped horizontally. For temporal augmentation, $64$ consecutive frames are picked randomly from the videos. Shorter videos are zero-padded on both sides to obtain $64$ frames. During testing, we use $224\times224$ center crops, apply the models over the full video, and average the predictions.

\subsubsection{EgoGesture Dataset}
\emph{EgoGesture dataset}~\cite{cao2017egocentric,zhang2018egogesture} is a large multimodal hand gesture dataset collected for the task of egocentric gesture recognition. This dataset contains $24,161$ hand gesture clips with $83$ gesture classes being performed by $50$ subjects. Videos in this dataset include both static and dynamic gestures captured with an Intel RealSense SR300 device in RGB-D modalities across multiple indoor/outdoor scenes.

We assess the performance of our method along with various hand gesture recognition methods published. Table~\ref{tbl:multi_ego} compares unimodal test accuracies for I3D on separate modalities and test accuracies of different hand gesture methods by fusion of RGB and depth. VGG16~\cite{simonyan2014very} processes each frame independently and VGG16+LSTM~\cite{donahue2015long} combines this method with a recurrent architecture to leverage the temporal information. As can be seen, the 3D CNN-based methods, C3D~\cite{tran2015learning}, C3D+LSTM+RSTMM~\cite{cao2017egocentric}, and I3D~\cite{carreira2017quo} outperform the VGG16-based methods. However, among the 3D CNN architectures, our method outperforms the top performers I3D late fusion by $0.73\%$.

\begin{table}[t]
\begin{center}
\resizebox{0.8\linewidth}{!}{%
\begin{tabular}{ l c c}
%\hlineB{3}
%\cline{3-7}
%\cline{2-3} 
Method   & Input Modalities & Accuracy\\% & Fusion (RGB-D)\\%   & Avg.          $\pm$  std.      \\
\hline 
I3D~\cite{carreira2017quo} &	RGB  & 90.33  \\
I3D~\cite{carreira2017quo} &	Depth  &  89.47  \\
\hline
VGG16~\cite{simonyan2014very} &	RGB+Depth & 	66.5		 \\
VGG16 + LSTM~\cite{donahue2015long} &	RGB+Depth	& 	81.4		 \\
C3D~\cite{tran2015learning} &	RGB+Depth & 	89.7		 \\
C3D+LSTM+RSTTM~\cite{cao2017egocentric} &	RGB+Depth & 	92.2		 \\
I3D late fusion~\cite{carreira2017quo} &	RGB+Depth 	& 	92.78		 \\
%MTUT~\cite{abavisani2019unimodal} &	RGB+Depth	& 	\bf{93.87}		 \\
Ours &	RGB+Depth	& 	\bf{93.51}		 \\

\hline\hline
%\hlineB{3}
\end{tabular}}
\caption{Accuracies of different multimodal fusion hand gesture methods on the EgoGesture dataset~\cite{cao2017egocentric}.}
%The top performer is denoted by boldface.} 
\label{tbl:multi_ego}
\end{center}
% \vspace{-2.5mm}
\vspace{-0.2in}
\end{table}

\subsubsection{NVGesture Dataset}
\label{sec:NVGesture}
\emph{NVGestures dataset}~\cite{molchanov2016online} was captured with multiple sensors for studying human-computer interfaces. It contains $1532$ dynamic hand gestures recorded from $20$ subjects inside a car simulator with artificial lighting conditions. This dataset includes $25$ classes of hand gestures. The gestures were recorded with SoftKinetic DS325 device as the RGB-D sensor and DUO-3D for the infrared streams. In addition, the optical flow and infrared disparity map modalities are usually used to enhance the prediction results. Following the previous works~\cite{molchanov2016online,Kopuklu_2018_CVPR_Workshops}, we only use RGB, depth, and optical flow modalities in our experiments. The optical flow is calculated using the method presented in~\cite{farneback2003two}. The RGB and optical flow modalities are well-aligned in this dataset, however, the depth map includes a larger field of view. % (see Figure~\ref{fig:datasets} (b)).  %Therefore, this experiment can examine the performance of our method in dealing with small misalignment across the modalities.

Table~\ref{tbl:multi_nv} presents the results of our method in comparison with the recent state-of-the-art methods: HOG+HOG2, improved dense trajectories (iDT)~\cite{wang2016robust}, R3DCNN~\cite{molchanov2016online}, two-stream CNNs~\cite{simonyan2014two} and MFFs~\cite{Kopuklu_2018_CVPR_Workshops}. We also report human labeling accuracy for comparison. The iDT~\cite{wang2016robust} method is often recognized as the best performing method with hand-engineered features~\cite{tran2018closer}. Similar to the previous experiment, we observe that the 3D-CNN-based methods outperform other hand gesture recognition methods, and among them, 
Our method provides the top performance in all the modalities. FOANet~\cite{Narayana_2018_CVPR} method achieves 91.28\% on this dataset using a sparse fusion method. However, this result is not comparable with the methods in Table~\ref{tbl:multi_nv} since FOANet relies on a separate pre-trained network to detect the hand. 

\begin{table}[t]
\small
\begin{center}
\resizebox{0.85\linewidth}{!}{%
\begin{tabular}{l c c}
%\hlineB{3}
%\cline{3-7}
Method & Input Modalities & Accuracy\\% & Fusion (RGB-D)\\%   & Avg.          $\pm$  std.      \\
\hline
I3D~\cite{carreira2017quo} &	RGB  & 78.42  \\
I3D~\cite{carreira2017quo} &	Opt. flow  & 83.19  \\
I3D~\cite{carreira2017quo} &	Depth  &  82.28 \\
\hline
HOG+HOG2~\cite{ohn2014hand} & RGB+Depth & 36.9  \\
I3D late fusion~\cite{carreira2017quo} &	RGB+Depth& 84.43  \\
%MTUT~\cite{abavisani2019unimodal} &	RGB+Depth	&  85.48  \\
Ours &	RGB+Depth	&  \bf{86.31}  \\
\hline
Two Stream CNNs~\cite{simonyan2014two} &	RGB+Opt. flow & 65.6  \\
iDT~\cite{wang2016robust} &	RGB+Opt. flow & 73.4 \\
R3DCNN~\cite{molchanov2016online} &	RGB+Opt. flow & 79.3		 \\
MFFs~\cite{Kopuklu_2018_CVPR_Workshops} &	RGB+Opt. flow & 84.7 \\
I3D late fusion~\cite{carreira2017quo} &	RGB+Opt. flow& 84.43  \\
%MTUT~\cite{abavisani2019unimodal} &	RGB+Opt. flow & \bf{85.48}\\
Ours &	RGB+Opt. flow &  \bf{84.85} \\

\hline
R3DCNN~\cite{molchanov2016online} &	RGB+Depth+Opt. flow & 83.8		 \\
I3D late fusion~\cite{carreira2017quo} &	RGB+Depth+Opt. flow  &85.68		 \\
%MTUT~\cite{abavisani2019unimodal} & RGB+Depth+Opt. flow & \bf{86.93}	 \\
Ours & RGB+Depth+Opt. flow & \bf{86.93}	 \\
\hline
Human~\cite{molchanov2016online} & &88.4 \\
\hline\hline
%\hlineB{3}
\end{tabular}}
\caption{Accuracies of different multimodal fusion hand gesture methods on the NVGesture dataset~\cite{molchanov2016online}.}
%The top performer is denoted by boldface.} 
\label{tbl:multi_nv}
\vspace{-0.2in}
\end{center}
\end{table}

\subsection{Audio-Visual Speech Enhancement}
\label{sec:av_experiment}
In this section, we evaluate our MMTM method on audio-visual speech enhancement. Using PESQ and STOI objective measures, we demonstrate that our slow fusion MMTM method outperforms state-of-the-art late fusion, channel-wise concatenation AVSE approaches. We use \emph{VoxCeleb2} \cite{chung2018voxceleb2}, a large audio-visual dataset obtained from YouTube that contains over 1 million utterances for 6,112 celebrities. The training, validation, and test datasets are split by celebrity ID (CID) such that the sets are disjoint over CIDs. In addition, CHiME-1/3 \cite{barker2013csl,Barker2017csl}, NonStationaryNoise \cite{duan2012online}, ESC50 \cite{piczak2015dataset}, HuCorpus \cite{hu2010tandem}, and private datasets are used for additive noise.

Video frames are extracted at $25$ FPS and S$^3$FD \cite{zhang2017s3fd} performs face detection. Following \cite{stafylakis2017combining}, we discard redundant visual information by cropping the mouth region via facial landmarks obtained from Facial Alignment Network \cite{bulat2017far}. Lip frames are resized to $122 \times 122$, transformed to grayscale, then normalized using the global mean and variance statistics from the training set. The audio waveform is extracted from the video following the methods of \cite{afouras2018conversation, gabbay2017visual}. We specify a window length of 40ms, hop size of 10ms, and sampling rate of 16kHz to align one video frame to four audio steps. Short-time Fourier transform (STFT) with a Hanning window function converts the waveform to spectrogram, $X_{spec} \in \mathbb{R}^{T \times F}$ with a frequency resolution of $F=321$, representing frequencies from $0-8$kHz. 

Training samples of batch size 4 are generated on-the-fly as lip frame and spectrograms pairs, ($X_{vid}, X_{spec}$). Interference spectrograms, $X_{inter}$, are sampled from the \emph{VoxCeleb2} set. We progressively increase the number of interference speakers during training, beginning with one and incrementing by one every 50 epochs until we reach the max of four. A noise spectrogram, $X_{n}$, is randomly sampled from the noise datasets. The mixture spectrogram is constructed via $X_{mix} = X_{spec} + \alpha X_{inter} + \beta X_{n}$, where $\alpha, \beta$ are mixing coefficients that achieve a specific SNR. Training and test SNRs are sampled from 0-20dB and 2.5-17.5dB ranges, respectively. $X_{mix}$ is transformed to a log-mel representation, $\log X_{mel} \in \mathbb{R}^{T \times F}$, where $T=116$ and $F=80$. We augment lip frames, $X_{vid}$, via random cropping ($\pm$ 5 pixels) and left-right flips. Augmented frames are resized to $112 \times 112$ and fed into the visual network.

%\subsubsection{Implementation Details} 

% Experiment Results
% Nice table reference: https://inf.ethz.ch/personal/markusp/teaching/guides/guide-tables.pdf

\begin{table}[tp]
\small
\centering
\resizebox{0.75\linewidth}{!}{%
\begin{tabular}{lccc} %\toprule 
Method & Fusion Method & PESQ & STOI \\
\midrule
Target & - & 4.64 & 1.000 \\
Mixed  & - & 2.19 & 0.900 \\
\hline
AVSE \cite{afouras2018conversation}$^\dagger$ & CWC & 2.59 & 0.650\\
\hline
AO Baseline & - & 2.43 & 0.930\\
AV Baseline & CWC & 2.67 & 0.938\\
% Bottleneck
				
Ours & MMTM & \best{2.73} & \best{0.941} \\
% Ours & MMTM-3 & 2.74 & 0.939 \\
% Ours & MMTM-2 & 2.70 & 0.938 \\
% Ours & MMTM-1 & \best{2.79} & 0.940 \\
\hline\hline
%\bottomrule
\end{tabular}}
\caption{Speech enhancement evaluations on the \emph{VoxCeleb2} dataset \cite{chung2018voxceleb2} for 3 simultaneous speakers. CWC: Channel-wise concatenation. %The best performance is denoted by boldface. 
$\dagger$ for approximate reference only.
}
\label{table:AVSE_results}
\vspace{-0.2in}
\end{table}

%%% RESULTS
% --------------------------------------------------
Objective evaluation results are shown in Table \ref{table:AVSE_results}. We evaluate enhanced speech using the perceptual quality of speech quality (PESQ) \cite{recommendation2001perceptual} and the short-time objective intelligibility (STOI) \cite{taal2011algorithm}. The audio only (AO) model is trained without the visual network and establishes an AO speech enhancement baseline. The AV baseline model establishes a baseline for predominant AVSE approaches that perform late fusion via CWC of AV features. We closely aligned the fusion mechanism in our AV baseline model architecture to that of \cite{afouras2018conversation}, and we matched the sample generation and training procedure as best we could given the information available. We report on \cite{afouras2018conversation} for reference only. 

Our AVSE model outperforms the AO and AV baselines on both objective measures PESQ and STOI. We outperform the AO baseline by 0.3 PESQ and 0.01 in STOI, demonstrating that visual information improves speech enhancement performance. Further, we outperform the AV baseline with CWC fusion by 0.06 PESQ, indicating that MMTM via slow fusion affords the greatest performance improvement. Our model generalizes to speakers unseen during training since CID is disjoint across train/test sets.
% --------------------------------------------------------

\subsection{Action Recognition}
NTU-RGBD dataset~\cite{Shahroudy_2016_CVPR} is a well-known large scale multimodal dataset. It contains $56,880$ samples captured from $40$ subjects performing $60$ classes of activities at $80$ view-points. Each action clip includes up to two people on the RGB video as well as $25$ body joints on 3D coordinate space. We followed the cross-subject evaluation~\cite{Shahroudy_2016_CVPR} that splits the $40$ subjects into training and testing sets. To have a fair comparison with previous works, we only use RGB and pose (skeleton) modalities. The architecture details can be found in Section~\ref{sec:action_recognition_app}. We followed section~\ref{sec:hand_gesture_result} for training settings as well as RGB data preparation and augmentation. 

Table~\ref{tbl:ntu-rgbd} shows the result of our method in comparison with the recent state-of-the-art methods on NTU-RGBD dataset. The first part of the table shows our unimodal baselines with I3D on RGB and HCN~\cite{li2018co} on skeletons. We use 3D skeletons and follow the $32$ frame subsampling method from the original paper. For simplicity in the fusion mechanism, we implemented multi-person slow fusion method~\cite{li2018co}. Consequently, our reported accuracy on HCN is lower than the result in~\cite{li2018co}. The second part shows state-of-the-art methods specifically design for action recognition by integrating RGB and skeleton. Our proposed fusion method outperforms all the recent action recognition algorithms. To our knowledge this is a new state-of-the-art result for RGB+Pose on the NTU-RGBD dataset~\cite{Shahroudy_2016_CVPR}.
\begin{table}[t]
\small
\begin{center}
\resizebox{.95\linewidth}{!}{%
\begin{tabular}{ l c c}
%\hlineB{3}
%\cline{3-7}
 Method & Input Modalities & Accuracy\\% & Fusion (RGB-D)\\%   & Avg.          $\pm$  std.      \\
\hline
%HCN~\cite{li2018co} & &	Pose   & 85.24   \\
%ResNet3D &	& RGB  & 83.91   \\
HCN$_{\text{ours}}$ &	Pose   & 77.96   \\
I3D~\cite{carreira2017quo} &	RGB  & 89.25   \\
\hline
DSSCA - SSLM~\cite{shahroudy2017deep}  &	RGB+Pose & 74.86  \\
Bilinear Learning~\cite{Hu_2018_ECCV} &	RGB+Pose & 83.0 \\ %85.5 \\
2D/3D Multitask~\cite{luvizon20182d} &	RGB+Pose & 85.5 \\ %88.60  \\
PoseMap~\cite{liu2018recognizing} & RGB+Pose & 91.71 \\
Late Fusion (I3D + HCN$_{\text{ours}}$) &	RGB+Pose &  91.56  \\
Ours &	RGB+Pose &  \bf{91.99} \\
\hline\hline
%\hlineB{3}
\end{tabular}}
\caption{Accuracies of different multimodal fusion action recognition methods on the NTU-RGBD dataset~\cite{Shahroudy_2016_CVPR}.} % The top performer is denoted by boldface.} 
\label{tbl:ntu-rgbd_fusion}
\end{center}
\vspace{-0.2in}
\end{table}
%P{\'e}rez-R{\'u}a

Next, we use the recently released code %\footnote{\hyperlink{https://github.com/juanmanpr/mfas}{https://github.com/juanmanpr/mfas}} 
of~\cite{perez2019mfas} to compare several general purpose multimodal fusion algorithms on this dataset. We implement and train the proposed method within this framework. To have an identical setting with other methods, we use inflated Resnet-50~\cite{baradel2018glimpse} for video processing and the implementation of HCN~\cite{li2018co} provided in this framework for skeleton processing. Table~\ref{tbl:ntu-rgbd_fusion} illustrates the performance of these unimodal networks as well as different state-of-the-art multimodal fusion methods. MFAS~\cite{perez2019mfas} is an architecture search algorithm that leverages a sequential architecture exploration method to find an optimal fusion architecture. In addition to the two stream CNN~\cite{simonyan2014two}, which is a late fusion algorithm, we also report the results of two intermediate fusion algorithms Gated Multimodal Units (GMU)~\cite{arevalo2017gated} and CentralNet~\cite{vielzeuf2018centralnet}. Our method outperforms the state-of-the-art MFAS method without an extensive model search on this dataset. We believe this performance could be further improved by a comprehensive architecture tuning.

\begin{table}[h]
\small
\begin{center}
\resizebox{.95\linewidth}{!}{%
\begin{tabular}{ l c c}
 Method & Input Modalities & Accuracy\\
\hline
HCN~\cite{li2018co} &	Pose   & 85.24   \\
Infalated Resnet-50~\cite{baradel2018glimpse} & RGB  & 83.91   \\
\hline
Two Stream~\cite{simonyan2014two} & RGB+Pose & 88.60  \\
GMU~\cite{arevalo2017gated} & RGB+Pose &  85.80 \\ 
CentralNet~\cite{vielzeuf2018centralnet} & RGB+Pose & 89.36   \\
MFAS~\cite{perez2019mfas} & RGB+Pose & 90.04   \\
Ours &	RGB+Pose &  \bf{90.11} \\
\hline\hline
\end{tabular}}
\caption{Comparison of state-of-the-art multimodal fusion algorithms on the NTU-RGBD dataset~\cite{Shahroudy_2016_CVPR}. All methods use HCN and Infalated Resnet-50 backbone unimodal architectures.} \label{tbl:ntu-rgbd}
\end{center}
\vspace{-0.2in}
\end{table}

\subsection{Analysis of the Network}
\label{sec:study}
To understand the effects of some of our model choices, we explore the performance of some variations of our model on the NVGesture dataset~\cite{molchanov2016online}. In particular, we compare our fusion method with different architectures in the transfer layer. We also explore using a different number of transfer layers when all the implementation details are the same as RGB+Depth gesture recognition network described in Section~\ref{sec:NVGesture}.

Since the spatial dimensions are aligned in this problem, we can directly concatenate the convolutional features without squeezing them in the MMTM. In order to keep the spatial dimensions of these features across the module, we also need to change all the fully connected layers in MMTM to convolution layers with kernel size $1$. This ensures that the number of parameters remains the same. We refer to this approach as convolutional MMTM. %One may argue that MMTM misses the local feature information by squeezing the spatial dimensions, which should decrease the performance of MMTM for this task. However, we show below that this is not the case and aggregating local information into a global descriptor is critical to get the best performance. 
In addition, we also use a variation of the convolutional MMTM that utilizes a sum operation instead of the gating operation. This approach is closely related to residual learning~\cite{he2016deep} and has been proposed for multimodal fusion with aligned spatial dimensions~\cite{hazirbas2016fusenet}. Finally, we evaluate the performance of the original Squeeze and Excitation (SE) approach in which each unimodal stream uses self excitation to recalibate its own channel-wise features. The scores of these unimodal networks are fused by late fusion at the end. 

Table~\ref{tbl:study_alternate} compares the accuracy of these variations, as well as their FLOPS and number of parameters with the late fusion and MMTM. Surprisingly, the convolutional MMTM variations do not show any noticeable improvement over the late fusion method. This result highlights the importance of extracting information with global receptive field information in the squeeze unit. We also note that not using the squeeze blocks increase the number of FLOPS by about $5$ times. Finally, the result of self excitation approach with no intermediate fusion clearly shows that the most of performance gain in MMTM is due to the slow fusion of the modalities rather than pure squeeze and excitation method.

\begin{table}[t]
\begin{center}
\resizebox{\linewidth}{!}{%
\begin{tabular}{ l c c c}
%\hlineB{3}
%\cline{3-7}
%\cline{2-3} 
 Method & Accuracy & \#FLOPS & \#Parameters \\
\hline
 Early Fusion  &  78.84 & $247M$ & $12.3M$ \\
 Late Fusion  &  84.43 & $405M$ & $24.6M$ \\
 Convlutional MMTM &  84.43 & $25.24G$ & $31.6M$ \\
 Convlutional MMTM (with sum op.) &  84.65 & $25.24G$ & $31.6M$ \\
 SE~\cite{Hu_2018_CVPR} + Late Fusion &  85.06 & $472M$ & $31.6M$ \\
 MMTM & \best{86.31} & $472M$ & $31.6M$ \\

\hline
\end{tabular}}
\caption{Comparison of different MMTM architectures on the NVGesture dataset.}
\label{tbl:study_alternate}
\vspace{-0.2in}
\end{center}
\end{table}

\begin{figure}[h]
	\begin{center}
		\includegraphics[width=.7\linewidth]{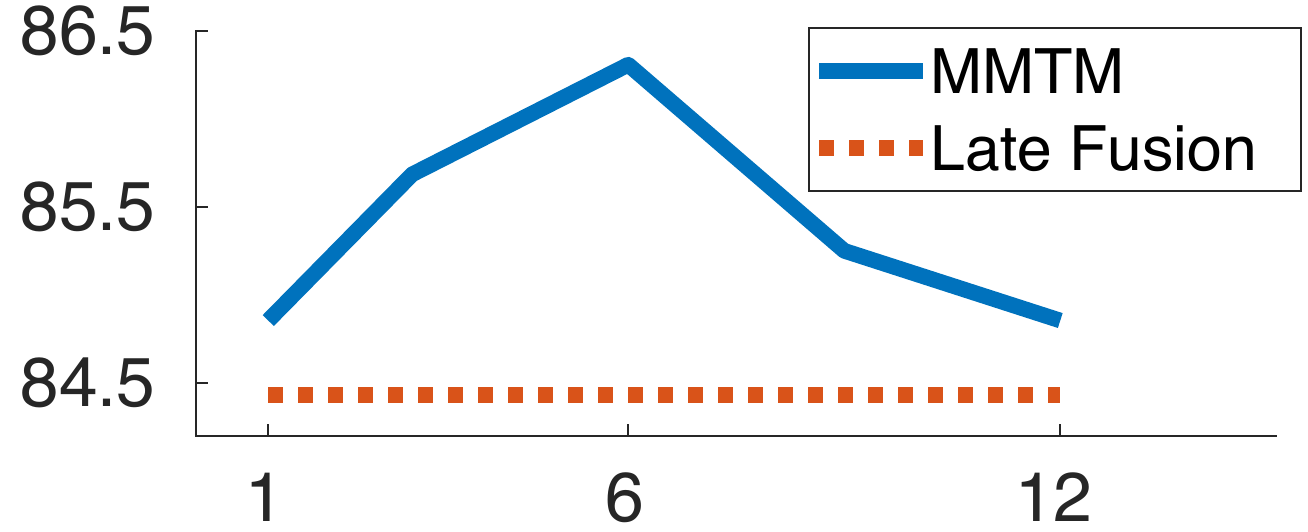}
	\end{center}
	\vspace{-0.1in}
	\caption{Accuracy vs. \#MMTMs on the NVGesture dataset.}
	\label{fig:study_number}
	\vspace{-0.2in}
\end{figure}

As we mentioned in Section~\ref{sec.app.visual}, we use MMTM after the last $6$ inception modules. In the last study, we evaluate the performance of the RGB+Depth gesture recognition network with MMTM applied to a different number of inception modules. Figure~\ref{fig:study_number} shows how the performance changes with respect to the number of MMTMs. This experiment indicates that the best performance is achieved when the output of half of the last inception modules ($6$ out of $12$) are fused by MMTM. This suggests that mid-level and high-level features benefit more than low-level features from this approach.

%------------------------------------------------------------------------
\section{Conclusion}
\label{sec:conclusion}
We present a simple neural network fusion module for leveraging the knowledge from multiple modalities in convolutional neural networks. The proposed module can be added at different levels of the feature hierarchy, allowing slow modality fusion. A wide range of experiments on applications with different types of modalities show applicability of the proposed module to gesture recognition, audio-visual speech enhancement, and human action recognition.

{\small
\setlength{\bibsep}{0pt}
\bibliographystyle{unsrtnat}
\bibliography{egbib,AVSE}
}

\end{document}